\documentclass[a4paper,twoside]{article}

\usepackage{epsfig}
\usepackage{subcaption}
\usepackage{calc}
\usepackage{amssymb}
\usepackage{amstext}
\usepackage{amsmath}
\usepackage{amsthm}
\usepackage{multicol}
\usepackage{pslatex}
\usepackage{apalike}
\usepackage{algorithm2e}
\usepackage[bottom]{footmisc}

\usepackage{tikz}
\usepackage{adjustbox}
\usepackage{xspace}
\newcommand{\circlemarker}[1]{%
  \begin{tikzpicture}[baseline=(char.base)]
    \shade[ball color=blue!15!cyan] (0,0) circle (0.16cm);
    \draw[line width=1pt, color=blue!80!cyan] (0,0) circle (0.16cm);
    \node[text=white, font=\bfseries] (char) at (0,0) {#1};
  \end{tikzpicture}%
}

\usepackage[acronym]{glossaries}
\glsdisablehyper
\newacronym{sl}{SL}{Supervised Learning}
\newacronym{ssl}{SSL}{Self-Supervised Learning}
\newacronym{cl}{contrastive learning}{Contrastive Learning}
\newacronym{cnn}{CNN}{Convolutional Neural Network}
\newacronym{vit}{ViT}{Vision Image Transformer}
\newacronym{dt}{DT}{Self-Distillation}
\newacronym{dl}{DL}{Deep Learning}
\newacronym{mvcnn}{MVCNN}{Multi-view Convolutional Neural Network}
\newacronym{mvtn}{MVTN}{Multi-view Transformation Network}
\newacronym{mvcl}{MVCL}{Multi-view Contrastive Learning}
\newacronym{cmvlnet}{CMVL-Net}{Contrastive Multi-view Learning Network}
\newacronym{map}{mAP}{mean Average Precision}
\newacronym{ce}{CE}{Cross-Entropy}

\usepackage{booktabs}
\usepackage{amsmath,amsthm,amssymb}
\usepackage{multirow}
\usepackage{xcolor}
\usepackage{hyperref}
\usepackage{diagbox}
\usepackage{slashbox}
\usepackage[table]{xcolor}
\usepackage{adjustbox}

\usepackage{bbding}
\definecolor{darkpastelgreen}{rgb}{0.01, 0.75, 0.24}
\def\yes{\textcolor{darkpastelgreen}{\Checkmark}\xspace}
\def\no{\textcolor{red!20}{\XSolidBrush}\xspace}

\def\best {\textcolor{green}{$\uparrow$}\xspace}

\usepackage{SCITEPRESS}     

\begin{document}

\title{Transformed Multi-view 3D Shape Features with Contrastive Learning}

\author{\authorname{Márcus Vinícius Lobo Costa\sup{1}\orcidAuthor{0000-0001-6727-1807}, 
Sherlon Almeida da Silva\sup{1}\orcidAuthor{0000-0001-6124-9350},  
Bárbara Caroline Benato\sup{1}\orcidAuthor{0000-0003-0806-3607}, \\
Leo Sampaio Ferraz Ribeiro\sup{1}\orcidAuthor{0000-0003-1781-2630},
and Moacir Antonelli Ponti\sup{1}\orcidAuthor{0000-0003-2059-9463}}
\affiliation{\sup{1}Instituto de Ciências Matemáticas e de Computação (ICMC), Universidade de São Paulo (USP), São Carlos, SP, Brazil}
\email{{marcusvlc, sherlon}@usp.br, barbarabenato@gmail.com, {leo.ribeiro, moacir}@icmc.usp.br}
}

\keywords{Representation Learning, Contrastive Learning Losses, Multi-view, Vision Transformers, Shape Understanding}

\abstract{This paper addresses the challenges in representation learning of 3D shape features by investigating state-of-the-art backbones paired with both contrastive supervised and self-supervised learning objectives. Computer vision methods struggle with recognizing 3D objects from 2D images, often requiring extensive labeled data and relying on Convolutional Neural Networks (CNNs) that may overlook crucial shape relationships.
Our work demonstrates that Vision Transformers (ViTs) based architectures, when paired with modern contrastive objectives, achieve promising results in multi-view 3D analysis on our downstream tasks, unifying contrastive and 3D shape understanding pipelines.
For example, supervised contrastive losses reached about 90.6\% accuracy on ModelNet10. The use of ViTs and contrastive learning, leveraging ViTs' ability to understand overall shapes and contrastive learning's effectiveness, overcomes the need for extensive labeled data and the limitations of CNNs in capturing crucial shape relationships.
The success stems from capturing global shape semantics via ViTs and refining local discriminative features through contrastive optimization. Importantly, our approach is empirical, as it is grounded on extensive experimental evaluation to validate the effectiveness of combining ViTs with contrastive objectives for 3D representation learning. The code is available at: \url{https://github.com/usmarcv/RepLearningLosses}.
}

\onecolumn \maketitle \normalsize \setcounter{footnote}{0} \vfill

\section{\uppercase{Introduction}}
\label{sec:introduction}

Understanding  3D shapes is crucial for applications such as robotics and virtual reality, yet it remains a challenging task for computer vision systems. While humans perceive 3D structure from 2D views, computational models still face significant limitations. Current methods often rely on multiple 2D projections of 3D objects processed by \acrfull{cnn} backbones trained under supervised learning. The representative work, Multi-View Convolutional Neural Network (MVCNN)~\cite{su2015multi}, made the first attempt at 3D shape understanding using CNN backbones. MVCNN was followed by further in-depth studies~\cite{hamdi2021mvtn,esteves2019equivariant}. However, these methods not only demand large labeled datasets but also struggle to capture complex shape relationships. In contrast, recent advances demonstrate that \acrfull{vit} architectures frequently outperform CNNs in visual recognition tasks~\cite{caron2021emerging,balestriero2023cookbook}, highlighting their potential for 3D understanding in multi-view learning settings.

In a scenario of crescent interest, self-supervised learning has emerged as a powerful paradigm for leveraging unlabeled data~\cite{balestriero2023cookbook}. By designing pretext tasks that generate pseudo-labels directly from the input, self-supervised learning showed to be more robust while mitigates the heavy reliance on manual annotation~\cite{cavallari2022training}, which remains a major bottleneck for supervised approaches~\cite{dos2020learning}.
While supervised learning continues to play a central role in many downstream applications, self-supervised learning offers a scalable alternative for data-efficient 3D representation learning.

In recent years, there has been a resurgence of works utilizing contrastive learning in many fields, particularly in settings with unlabeled examples. Contrastive learning operates in settings with both supervised and self-supervised learning with a central idea: ``\textit{pull together an anchor and a positive sample in an embedding space, and push apart the anchor from many negative samples}''~\cite{oord2018infonce,khosla2020supcon}. These works employ InfoNCE~\cite{oord2018infonce} and SimCLR~\cite{Chen2020simclr} loss functions in supervised contrastive settings, such as SupCon~\cite{khosla2020supcon} loss. 

Taking advantage of these approaches, some progress has been made in 3D shape understanding through contrastive learning in self-supervised settings~\cite{li2025multi,peng2024contrastive}. CMVL-Net~\cite{peng2024contrastive} utilizes embeddings generated by a CNN that relies on contrastive clustering via graph methods, while MVCL~\cite{li2025multi} uses a discriminative multi-view grouping that learns correlation between them using contrastive learning through a CNN backbone. However, contrastive learning settings have been considered only when combined with CNN backbones for 3D multi-view shape understanding.

In this paper, we investigate how to integrate and leverage \textit{contrastive learning and ViT for 3D shape understanding with multi-view rendering}. We propose to address the abovementioned limitations by exploring in particular four ViTs backbones and five contrastive learning losses, including self-supervised and supervised settings. We compare our proposed experiments with approaches using CNN backbones to evaluate what works best for 3D shape understanding in an end-to-end pipeline. 
We conduct our experiments in a way to answer the following research questions (RQ):
\begin{description}
    \item[\textbf{RQ1.}] \textbf{How CNN and ViT backbones perform for multi-viewing 3D shape?} Following the findings in~\cite{Chen2020simclr,khosla2020supcon,li2025multi} showing that ResNet-50~\cite{he2016deep} show beneficial performance in the classification task. In addition, we explored the ViT variants as ViT-S/16~\cite{dosovitskiy2020image}, ViT-B/16, DINO-S/16~\cite{caron2021emerging} and DINO-B/16.

    \item[\textbf{RQ2.}] \textbf{How contrastive learning losses can enhance (or contribute to) multi-view 3D rendering pipelines?} Besides the cross-entropy loss, we explored five alternatives to contrastive learning losses, such as InfoNCE~\cite{oord2018infonce}, SimCLR~\cite{Chen2020simclr}, SupCon~\cite{khosla2020supcon}, $\epsilon$-SupInfoNCE~\cite{barbano2023unbiased}, and SINCERE~\cite{Feeney2023sincere}.
    
    \item[\textbf{RQ3.}] \textbf{How combining contrastive learning losses with ViT for multi-view 3D shape can improve state-of-art methods?}
    
\end{description}

With these research questions, our experimental setup can be summarized by the model variants obtained from the combination of the choices listed in response to the research questions. Our set of contributions is then multi-faceted:

\begin{itemize}
    \item we show ViT backbones overcome CNN ones with contrastive learning losses for 3D multi-view images, an underexplored combination for representation learning in 3D shape understanding.

    \item we present a comprehensive evaluation unifying the contrastive learning and 3D shape understanding pipelines.
    \item in-depth analysis of feature learning breadth of side-by-side comparisons, with visualization, analysis of limitations of the learned embeddings, and generalization assessment with evaluation on image retrieval tasks.
\end{itemize}

\noindent
\textbf{Outline.}
The paper is organized as follows. In Section \ref{sec:related}, we review deep learning, supervised and self-supervised learning, and contrastive learning under the multi-view 3D data. 
In Section \ref{sec:proposed-approach}, we present our proposed approach that unifies the contrastive learning and 3D shape understanding pipelines used in this work. 
Then, in Section \ref{sec:experiments-eval-discussion} we present the experimental protocol. 
In Section~\ref{sec:results}, we show the results obtained for our downstream tasks.
Finally, in Section \ref{sec:conclusions}, we point out the contributions, limitations, and future work.

\section{\uppercase{Related Works}}
\label{sec:related}

\noindent
\textit{\textbf{CNNs on 3D Multi-view data.}}
Deep learning methods for 3D object representation typically fall into three categories: voxel-based, point-based, and view-based approaches~\cite{qi2021review}. Looking to only view-based projections, the pioneering \acrfull{mvcnn}~\cite{su2015multi} introduced view-based methods with a unified \acrshort{cnn} architecture, employing view pooling for classification and retrieval tasks. Subsequent works advanced viewpoint optimization: \acrfull{mvtn}~\cite{hamdi2021mvtn} proposed a differentiable module to predict optimal camera viewpoints, while \cite{esteves2019equivariant} introduced the equivariant multi-view network (EMVN), aggregating views via polyhedron-based representations (e.g., icosahedrons) and ResNet-18 backbones for classification and retrieval. However, in this paper, we propose utilizing the ViT variants as backbones along with the CNN.

\noindent
\textit{\textbf{Supervised learning and Self-supervised learning.}}
Self-supervised learning (\acrshort{ssl}) has emerged as a powerful paradigm for leveraging unlabeled data, avoiding the cost and effort of manual annotation required in \acrfull{sl}~\cite{balestriero2023cookbook,dos2020learning}. Contrastive learning, a key \acrshort{ssl} approach, learns discriminative features by attracting similar instances and repelling dissimilar ones in the latent space~\cite{le2020contrastive,bui2018sketching}. While recent works, such as SupCon~\cite{khosla2020supcon}, $\epsilon$-SupInfoNCE~\cite{barbano2023unbiased}, and SINCERE~\cite{Feeney2023sincere}, explore hybrid \acrshort{sl}/\acrshort{ssl} losses in contrastive learning pipelines, their application to 3D understanding pipeline data with modern backbones remains underexplored: a gap our work addresses.

\noindent
\textit{\textbf{Contrastive learning on 3D data.}}
Recent efforts like \acrfull{mvcl}~\cite{li2025multi} use multi-view grouping to enhance feature discrimination, while \acrfull{cmvlnet}~\cite{peng2024contrastive} integrates clustering into contrastive learning. However, these methods rely on \acrshort{cnn} backbones or lack systematic evaluation across retrieval tasks. 
In contrast, we integrated ViT-based architectures with a contrastive learning pipeline for 3D multi-view data, enabling scalable representation learning and rigorous benchmarking on shape/image retrieval.

\noindent
\textit{\textbf{Summary of related works.}} 
Table~\ref{tab:summary-relatedworks} systematically compares prior works across five critical dimensions: contrastive learning settings, ViT backbones, $k$-NN classification, shape retrieval, and 3D shape understanding pipeline. 
While existing methods achieve partial coverage of such experimental setups (e.g., \acrshort{mvcnn} and \acrshort{mvtn} focus only on shape retrieval with \acrshort{cnn}s; \acrshort{mvcl} and CMVL-Net introduces contrastive learning but omits ViTs and 3D retrieval tasks), our work is the first to unify all five aspects, as evidenced by the table’s bottom row.

\begin{table}[htb]
\centering
\renewcommand*{\arraystretch}{1.2}
\setlength{\tabcolsep}{2.0pt}
\caption{Summary of related work.}
\label{tab:summary-relatedworks}
\resizebox{\columnwidth}{!}{%
\begin{tabular}{lccccc}
\toprule

  Work &
  \multicolumn{1}{c}{\begin{tabular}[c]{@{}c@{}}Contrastive\\Pipeline\end{tabular}} &
  \multicolumn{1}{c}{\begin{tabular}[c]{@{}c@{}}ViT\\Backbones\end{tabular}} &
  \multicolumn{1}{c}{$k$-NN} &
  \multicolumn{1}{c}{\begin{tabular}[c]{@{}c@{}}3D Shape\\Retrieval\end{tabular}} &
  \multicolumn{1}{c}{\begin{tabular}[c]{@{}c@{}}3D \\ Pipeline\end{tabular}}
  
  \\ 
  
\midrule

\acrshort{mvcnn} \cite{su2015multi}                 &   \no &  \no &  \no & \yes  &  \yes \\

\acrshort{mvtn} \cite{hamdi2021mvtn}                &   \no &  \no &  \no & \yes  &  \yes \\
                                        
EMVN \cite{esteves2019equivariant}                  &   \no &  \no  &  \no & \no  &  \no \\

\midrule

\acrshort{mvcl} \cite{li2025multi}                  &   \yes &  \no & \no & \no  &  \no \\

\acrshort{cmvlnet} \cite{peng2024contrastive}       &   \yes &  \no & \yes & \yes &  \no \\


\midrule

\rowcolor{orange!10} 
\textbf{Ours}    & \yes & \yes & \yes & \yes & \yes \\ 

\bottomrule

\end{tabular}%
}
\smallskip
\begin{minipage}{\columnwidth}
\scriptsize{\smallskip The ``\yes'' means the work matches the aspects, and ``\no'' does not match the aspect specifications.}
\end{minipage}
\end{table}

\section{\uppercase{Proposed Approach}}
\label{sec:proposed-approach}

\begin{figure*}[htb]
    \centering
    \includegraphics[width=.99\linewidth]{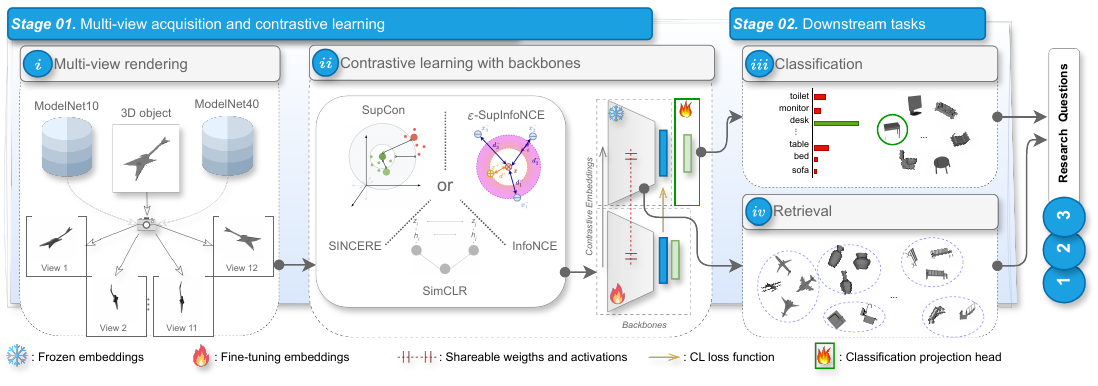}
    \caption{Overview of the experimental pipeline, consisting of two stages. Stage 01 performs multi-view rendering from 3D objects (e.g., ModelNet10/40), followed by contrastive learning using different loss functions (e.g., SupCon, InfoNCE, SimCLR, SINCERE, and $\epsilon$-SupInfoNCE) and backbones. Embeddings can be frozen or fine-tuned. Stage 02 evaluates the learned representations through downstream tasks such as 3D shape classification and retrieval. The entire pipeline is designed to answer our three main research questions.}
    \label{fig:workflow}
\end{figure*}

As discussed above, we explore empirically different contrastive learning losses with view-based descriptors for 3D shape objects. Our approach consists of four key components: $(i)$ multi-view rendering, $(ii)$ contrastive learning with backbones, $(iii)$ classification, and $(iv)$ retrieval. Figure~\ref{fig:workflow} illustrates the complete pipeline for our scenario using 3D multi-view shape features.

\subsection{Multi-view acquisition and contrastive learning (Stage 01)}
In this stage, given a 3D shape dataset with multi-view rendering images, we aim to:
\begin{itemize}
    \item Learn discriminative representations using contrastive learning losses with both ViT and CNN backbones;
    \item Benchmark performance of diverse contrastive losses (e.g., InfoNCE, SimCLR, SupCon, $\epsilon$-SupInfoNCE, and SINCERE).
\end{itemize}

\noindent \textbf{}\textbf{\textit{i) Multi-view Rendering.}}
Open-source 3D datasets typically provide objects represented as meshes $\mathcal{M}$, which are discrete approximations of continuous surfaces $S \subset \mathbb{R}^3$. A mesh can be formally defined as a collection of vertices $V = \{ \mathbf{v}_i \in \mathbb{R}^3 \mid i = 1, \dots, N_v \}$, connected by edges $E = \{ (i, j) \mid \mathbf{v}_i, \mathbf{v}_j \in V \}$, that together form triangular faces $F = \{ (i, j, k) \mid (i, j), (j, k), (k, i) \in E \}$.

\begin{figure}[htb]
    \centering
    \includegraphics[width=.99\linewidth]{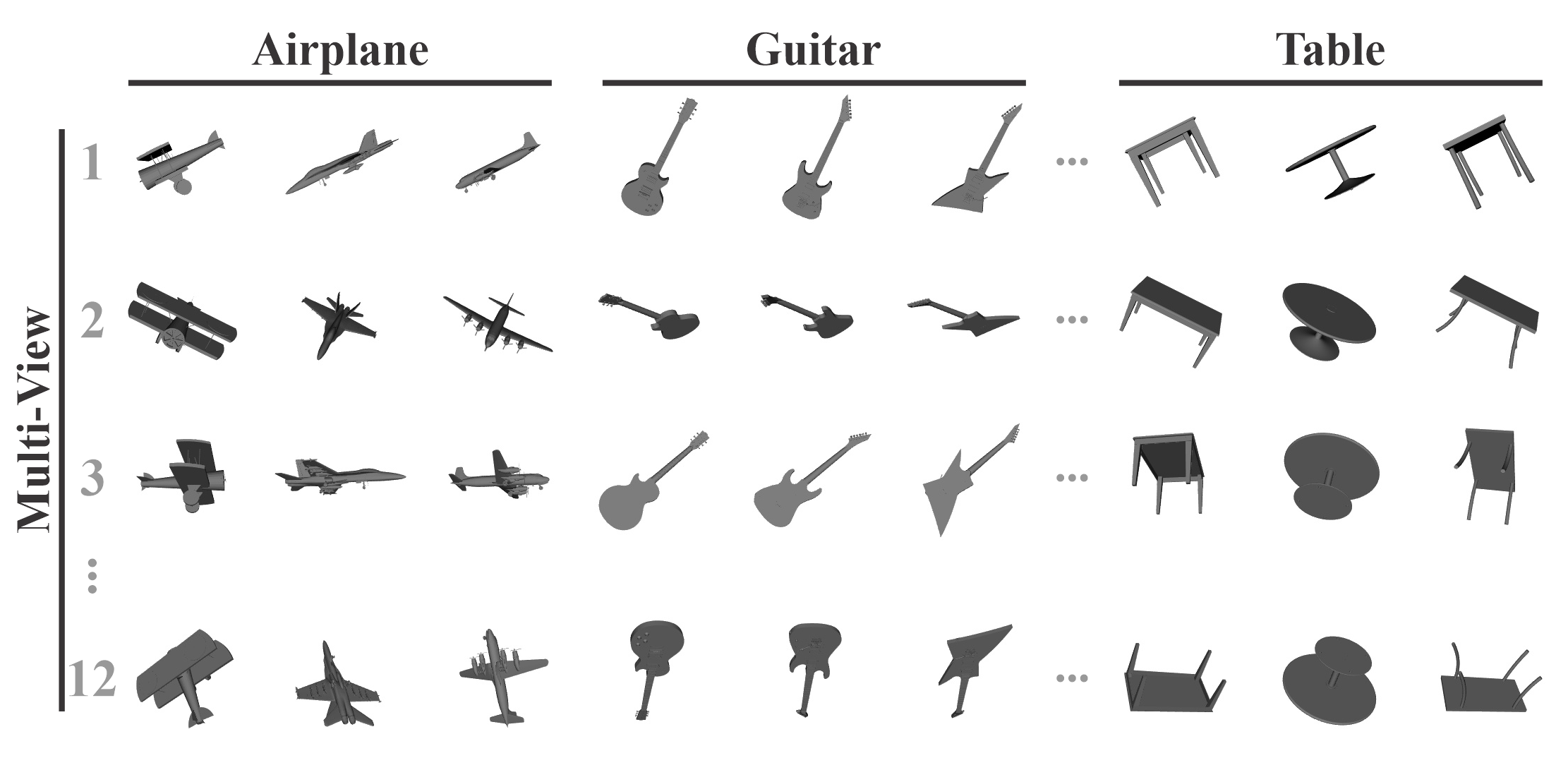}
    \caption{Multi-view rendering acquisition from 3D objects relies on our camera setup. Some representative examples from the airplane, guitar, and table classes to illustrate the diversity of views generated.}
    \label{fig:multiview_example}
\end{figure}

From these 3D meshes, we automate the rendering of multi-view images using Python~3.11 and the VEDO library~\cite{musy2021vedo}\footnote{VEDO: \url{https://vedo.embl.es/}}. The image acquisition process begins by loading a 3D object and applying a sequence of controlled rotations to align and center the mesh relative to the camera. Following the approach of~\cite{su2015multi}, the mesh polygons are rendered under a perspective projection, where each pixel color is obtained by interpolating the intensities reflected from the polygon vertices. Finally, we generate 2D images by performing multi-view rendering of each 3D mesh using 12 consecutive rotations around the $(X, Y, Z)$ axes. Although the output resolution is configurable, we fix it at 224$\times$224 pixels to ensure compatibility with widely adopted state-of-the-art architectures. Figure~\ref{fig:multiview_example} illustrates several examples of the rendered multi-view images.

\vspace{0.1cm}
\noindent \textbf{}\textbf{\textit{ii) Contrastive learning with backbones.}}
Inspired by recent contrastive learning losses (e.g., SimCLR~\cite{Chen2020simclr}, and SupCon~\cite{khosla2020supcon}), our approach learns discriminative representations by maximizing agreement between two augmented views of the same 3D object through a contrastive loss in the latent space. We implement both self-supervised and supervised contrastive learning approaches, in addition to the baseline based on CE loss. From self-supervised learning methods, we explore SimCLR~\cite{Chen2020simclr} and InfoNCE~\cite{oord2018infonce}. Besides self-supervised learning, we explore modern contrastive supervised learning methods, including SupCon~\cite{khosla2020supcon}, $\epsilon$-SupInfoNCE~\cite{barbano2023unbiased}, and SINCERE~\cite{Feeney2023sincere}. All these contrastive losses are experimented with in backbones based on CNN and ViT with variants. 

Our approach is composed of the following components:
\begin{itemize}
    \item A \textit{data augmentation} module, $Aug(\cdot)$. For each input sample from a view of a 3D object, $x$, we generate two random augmentations, $\tilde{x}=Aug(x)$. These augmentations offer a different perspective on the 3D data and include a representative subset of information from the original sample view.

    \item A backbone \textit{encoder network}, ${Enc(\cdot)}$. That maps $x$ and extracts its corresponding representation vector embeddings from augmented data examples. Our proposed approach allows backbones to rely on CNN and ViT variants.

    \item A small \textit{projection network}, $Proj(\cdot)$. That produces normalized embeddings for the representation vectors to be fed to the loss function. The $Proj(\cdot)$ is either a multi-layer perceptron~\cite{Trevor2001elements} with a single hidden layer and an output single linear layer.
\end{itemize}

In summary, the training process consists of two stages:
\begin{enumerate}
    \item \textbf{\textit{Pre-training}}: Apply the augmentations ($Aug(\cdot)$) with contrastive loss to train the backbone encoder (${Enc(\cdot)}$);
    \item \textbf{\textit{Linear evaluation}}: Discard the projection head ($Proj(\cdot)$) and train the linear evaluation protocol on frozen embeddings using cross-entropy loss.
\end{enumerate}

As mentioned above, previous work on \acrshort{cl} opt to use only backbones based on \acrshort{cnn}; in our work, we adapt these losses to perform in \acrshort{vit} backbones.
At the inference time with linear classification, the contrastive projection head is discarded. 
%

\subsection{Downstream tasks (Stage 02)}
We claim that our proposed approach, utilizing contrastive learning with 3D multi-view representation, can be applied to various types of tasks. We aim to:
\begin{itemize}
    \item Evaluate representations in classification ($k$-NN and linear evaluation);
    \item Evaluate 3D shape retrieval.
\end{itemize}

\noindent \textbf{}\textbf{\textit{iii) Classification.}}
We evaluate representations through:
\begin{itemize}
    \item Linear evaluation trained on frozen embeddings;
    \item $k$-NN classification (no fine-tuning);
    \item t-SNE \cite{van2008visualizing} visualization of learned embeddings.
\end{itemize}
To utilize the contrastive embedding representation, a linear classifier is appended to the top of the frozen representations, using a cross-entropy loss. 

\vspace{0.1cm}
\noindent \textbf{}\textbf{\textit{iv) Retrieval.}}
Our shape retrieval protocol:
\begin{itemize}
    \item \textbf{Query}: Given a query shape $S_q$, find the most similiar shapes in a broader set o size $N$;
    \item \textbf{Retrieval}: Rank shapes by cosine similarity of $l_2$-normalized embeddings;
    \item \textbf{Evaluation}: Use training set as retrieval corpus; test on held-out queries.
\end{itemize}

\section{\uppercase{Experimental setup}}
\label{sec:experiments-eval-discussion}
\label{subsec:exp-setup}
\label{sec:experimental_setup}
Our experimental setup is as follows: Section~\ref{subsec:exp-setup} outlines the datasets, contrastive learning losses, backbones, implementation details, and our evaluation metrics. 

\vspace{0.1cm}
\noindent
\textit{\textbf{Datasets.} }
To evaluate our downstream tasks, we used the Princeton ModelNet Dataset\footnote{Princeton ModelNet: \url{https://modelnet.cs.princeton.edu/}}.
ModelNet40~\cite{wu20153d} is a synthetic collection of 12,311 3D objects from 40 common categories. Similarly, ModelNet10~\cite{wu20153d} is a subset that contains only 10 representative categories. In addition, we utilize the training and testing split provided by the authors (80/20\% for train and test, respectively).

\vspace{0.1cm}
\noindent
\textit{\textbf{Backbones.}}
We select ResNet-50~\cite{he2016deep}, ViT-S/16~\cite{dosovitskiy2020image}, ViT-B/16\cite{dosovitskiy2020image}, DINO-S/16~\cite{caron2021emerging} and DINO-B/16~\cite{caron2021emerging}. All weights were initialized with ImageNet~\cite{russakovsky2015imagenet}\footnote{ImageNet: \url{https://www.image-net.org/}} pre-training.
Our backbones are structurally similar to those used in~\cite{Chen2020simclr,khosla2020supcon} with ResNet-50, which includes an encoder with 2048-dimensional normalized embeddings, followed by a linear projection head with 128-dimensional embeddings. To backbone-based on \acrshort{vit} architectures, we explore small and base variants. The small variant comprises 12 layers, 6 attention heads, and a linear projection head at the output, producing 384-dimensional embeddings. For the base variant, 12 layers, 12 attention heads, and a linear projection head with 768-dimensional embeddings are used. In contrast, the DINO~\cite{caron2021emerging} architecture is based on the \acrshort{vit} backbone but is trained under a self-distillation protocol.
We used the same variants applied under DINO.

\vspace{0.1cm}

\noindent \textit{\textbf{Contrastive learning losses.}} To evaluate the pipeline for contrastive learning in 3D shape understanding, we explore contrastive SSL losses, as: 
\begin{itemize}
    \item InfoNCE~\cite{oord2018infonce}: Maximizes mutual information between positive pairs;
    \item SimCLR~\cite{Chen2020simclr}: Uses augmentations to create positive pairs.
\end{itemize}

As contrastive SL losses we included:
\begin{itemize}
    \item SupCon~\cite{khosla2020supcon}: Pulls together same-class samples while pushing apart different classes;
    \item SINCERE~\cite{Feeney2023sincere}: Eliminates intra-class repulsion present in SupCon;
    \item $\epsilon$-SupInfoNCE~\cite{barbano2023unbiased}: Provides precise control over positive/negative sample distances.
\end{itemize}
\vspace{0.1cm}

\noindent
\textit{\textbf{Implementation details.}}
We conducted all experiments using a single NVIDIA A40 48 GB VRAM. We used PyTorch\footnote{PyTorch: \url{https://pytorch.org/}} 2.6 with CUDA 12.4.
All of our backbones are trained with a batch size of 128 for 100 epochs and a temperature $\tau$ = 0.07. The SGD optimizer with a learning rate of $1e-4$ with an exponential learning rate schedule with weight decay of $1e-4$. To accelerate and reduce GPU usage without compromising performance, we used \textit{automatic mixed precision}\footnote{Automatic Mixed Precision: \url{https://docs.pytorch.org/docs/stable/amp}} (with float16).
Our data augmentation is based on SimCLR \cite{Chen2020simclr} and SupCon \cite{khosla2020supcon}, which generate two augmented views of the same image using a sequence of transformations, including random resized cropping, horizontal flipping, color jittering, random grayscale conversion, and normalization. These augmentations introduce appearance variability while preserving semantic content, encouraging the model to learn invariant and discriminative features across different views of the same instance.
For $\epsilon$-SupInfoNCE, the parameter $\epsilon$ chosen was of 0.25 following~\cite{barbano2023unbiased}.
Such hyperparameters were chosen after an extensive hyperparameter search using $5$-fold cross-validation search within each training set.

\vspace{0.1cm}
\noindent
\textit{\textbf{Metrics.}}
We evaluate our performance by measuring the classification Accuracy (top-1 and top-5) and the Nearest Neighbor for 3D shape classification. For retrieval tasks, it is evaluated by \acrfull{map} over the test queries. 

\section{\uppercase{Results}}
\label{sec:results}

This section presents our experimental evaluation across our downstream tasks: 3D baseline classification, 3D shape classification, and 3D retrieval. 
First, in Section~\ref{subsec:3d-baseline} we compare our performance with cross-entropy loss through our CNN and ViT backbones. The contrastive learning and 3D pipeline performance are introduced in Section~\ref{subsec:contrastive}. 
In detail, we explore 3D shape classification under contrastive settings in Section~\ref{subsec:3d-contrastive-class}. The retrieval performance and shape retrieval tasks are in Section \ref{subsec:3d-retrieval}. In Section \ref{subsec:results-related}, we compare our performance with the literature works. Finally, in Section \ref{subsec:discussions}, we discuss the performance of our results.

The main results of our approach are summarized in Tables~\ref{tab:lineareval-acc-ce},~\ref{tab:lineareval-acc}, and~\ref{tab:all-metrics}. We achieve SOTA performance in 3D classification on ModelNet10 (90.6\%) and achieve the best $k$-NN performance (87.0\%) on ModelNet40. On shape retrieval, we achieve the best mAP performance (95.5\%) on ModelNet10. To follow a common practice, we report the best results out of one run in benchmark tables, but more detailed results and the code are available in our GitHub repository\footnote{Repository: \url{https://github.com/usmarcv/RepLearningLosses}}.

\subsection{RQ1: 3D Shape Classification}
\label{subsec:3d-baseline}

In this experiment, we revisit RQ1 (see Sec.~\ref{sec:introduction}). To answer this question, we propose to compare features provided from CNN-based and ViT-based backbones. In this step,~\textit{our main goal is not to evaluate contrastive learning scenarios yet, but only to evaluate and compare features provided from such backbones.} 
For this, we utilize both datasets for training and testing samples with a common Cross-Entropy (CE) loss updated via backpropagation. We present \emph{Top-1} and \emph{Top-5} accuracies over the test set.

Table~\ref{tab:lineareval-acc-ce} compares the performance of linear evaluation end-to-end between all choose backbones using CE loss over multi-view 3D shape images. The top-1 classification accuracy of DINO-S/16 in both datasets surpasses that of the ResNet-50 architecture by about 44\% \best. Following the same comparison to the top-5 classification accuracy for both datasets, with improvements ranging from about 15\% \best (ModelNet10) to 24\% \best (ModelNet40). The above results demonstrate that ViT variants can effectively learn discriminative representations based on specific architectures derived from ViT. ViT showed improved classification performance when compared with a CNN backbone -- the one most used in the literature for such applications.

\begin{table}[thb]
\centering
\renewcommand*{\arraystretch}{1.2}
\setlength{\tabcolsep}{4.0pt}
\caption{Backbones performance Top-1 and Top-5 accuracy (\%) on the test set using cross-entropy loss. Best in \textbf{bold}.}
\label{tab:lineareval-acc-ce}
\begin{adjustbox}{width=0.8\linewidth}
\begin{tabular}{llcc}
        \toprule
        \multicolumn{1}{l}{Dataset} & \multicolumn{1}{l}{Backbone} 
        & Top-1 (\%) & Top-5 (\%) \\
        \midrule
        \multirow{5}{*}{ModelNet10} 
          & ResNet-50   & 37.57 & 82.34 \\
          & ViT-S/16    & 44.61 & 85.06 \\
          & ViT-B/16    & 58.66 & 94.15 \\
          & DINO-S/16   & \textbf{81.87} & \textbf{98.28} \\
          & DINO-B/16   & 54.59 & 92.06 \\
        \cmidrule(lr){1-4}
        \multirow{5}{*}{ModelNet40} 
          & ResNet-50   & 32.57 & 68.99 \\
          & ViT-S/16    & 33.60 & 68.49 \\
          & ViT-B/16    & 55.45 & 84.72 \\
          & DINO-S/16   & \textbf{78.85} & \textbf{93.76} \\
          & DINO-B/16   & 55.64 & 85.83 \\
        \bottomrule
\end{tabular}
\end{adjustbox}
\end{table}

        

\subsection{RQ2: Contrastive Learning for Classification and Retrieval}
\label{subsec:contrastive}
We revisit RQ2 (see Sec.~\ref{sec:introduction}) to evaluate our proposed approach (Sec.~\ref{sec:proposed-approach}), with Stages 1 and 2. Our goal is to show the ability of contrastive learning losses in improving 3D multi-view with 2D rendering pipelines: (a) for classification (Sec.~\ref{subsec:3d-contrastive-class}) and (b) for retrieval tasks (Sec.~\ref{subsec:3d-retrieval}). To achieve this, we now explore CNN and ViT backbones trained with contrastive losses under supervised and self-supervised settings. 
For this, we train and test each backbone with the selected contrastive learning losses (see Sec.~\ref{sec:experimental_setup}).

\subsubsection{3D Contrastive Learning Shape Classification}
\label{subsec:3d-contrastive-class}
We compare contrastive learning results with CE loss results presented in last experiment (Sec.\ref{subsec:3d-baseline}). We present~\emph{Top-1} and~\emph{Top-5} accuracies over the test set. Table~\ref{tab:lineareval-acc} reports the classification accuracy of five losses and the CE loss (as baseline), for two datasets, and five different backbones for 3D multi-view images.

Several interesting results stem from this experiments. First, we observe that all proposed combinations of backbones and contrastive learning losses obtained higher performances when compared with baseline performance (CE loss) achieved higher performance compared to the ResNet-50, and InfoNCE and SimCLR losses. This, in fact, confirms our claim of using ViT instead of CNN as backbone for this application.

Now, considering only \emph{Top-1} results. For ModelNet10, we see that performances of self-supervised losses (InfoNCE and SimCLR) are around 40-44\% for CNN backbone and around 70-79\% with for ViT backbones. Also, we see that performances of supervised losses are around 78-80\% for CNN backbone and 83-91\% for ViT backbones. This confirms (i) the benefit of combining ViT and contrastive losses and (ii) the superior ability of supervised losses compared to self-supervised even for ViT backbones (83-91\% and 70-79\%, respectively) for multi-viewing 3D shape applications. For ModelNet40, performances of self-supervised losses are around 29-33\% for CNN backbone and around 66-76\%. Also, performances of supervised losses are around 56-58\% for CNN backbone and around 80-86\% for ViT backbones. Again, this confirms the benefit of using such combination of contrastive losses with combination ViT backbones over CNN backbones.

Considering \emph{Top-5} results in the following. For ModelNet10, performances of self-supervised losses are around 82-86\% for CNN backbone and around 85-99\% with for backbones. Performances of supervised losses are around 90-91\% for CNN backbone and 98-100\% for ViT backbones. Again, our results confirm (i) the benefit of combining ViT and contrastive losses. Related to (ii) the superior ability of supervised losses compared to self-supervised, for ViT backbones results are really close depending of the ViT architecture (85-99\% and 98-100\%, respectively). For ModelNet40, performances of self-supervised losses are around 75-76\% for CNN backbone and around 60-95\%. Performances of supervised losses are around 83-86\% for CNN backbone and around 73-98\% for ViT backbones. Again, this confirms the benefit of using such combination of contrastive losses with ViT backbones over CNN backbones, mainly ViT architectures as DINO-S/16 and DINO-B/16.

\begin{table*}[htb]
\centering
\renewcommand*{\arraystretch}{1.2}
\caption{Linear evaluation on Test Set. Performance evaluation Top-1 and Top-5 accuracy (\%) on the test set using various backbones and contrastive loss functions with supervised and self-supervised settings. \textbf{Bold} indicates the top-performing supervised contrastive loss per backbone, and \underline{underline} indicates the top-performing self-supervised contrastive loss.}
\label{tab:lineareval-acc}
\resizebox{\textwidth}{!}{%
\begin{tabular}{llcccccccccc}
        \toprule
        & & \multicolumn{5}{c}{Top-1 (\%)} & \multicolumn{5}{c}{Top-5 (\%)} \\
        \cmidrule(lr){3-7} \cmidrule(lr){8-12}
        
        \multicolumn{1}{l}{Dataset} & \multicolumn{1}{l}{Losses} 
        & ResNet-50 & ViT-S/16 & ViT-B/16 & DINO-S/16 & DINO-B/16 
        & ResNet-50 & ViT-S/16 & ViT-B/16 & DINO-S/16 & DINO-B/16 \\
        \midrule
        \multirow{6}{*}{ModelNet10} 
        & Cross-Entropy (baseline) & 37.57 & 44.61 & 58.66 & 81.87 & 54.59 & 82.34 & 85.06 & 94.15 & 98.28 & 92.06 \\
        \cmidrule(lr){2-12}
        
        & InfoNCE~\cite{oord2018infonce} & 41.99 & 70.93 & \underline{71.25} & 75.38 & 78.39 
                                         & 83.58 & \underline{97.59} & \underline{97.67} & 98.25 & 98.56 \\
        & SimCLR~\cite{Chen2020simclr} & \underline{43.62} & \underline{71.40} & 70.64 & \underline{76.49} & \underline{78.82} 
                                       & \underline{85.18} & 97.49 & 97.65 & \underline{98.41} & \underline{98.67} \\

         \cmidrule(lr){2-12}
        
        & SupCon~\cite{khosla2020supcon} & \textbf{80.08} & 83.24 & 90.49 & 90.16 & 90.14 
                                         & \textbf{98.02} & 98.84 & 99.67 & 99.48 & 99.59 \\
        & $\epsilon$-SupInfoNCE~\cite{barbano2023unbiased} & 77.10 & 89.71 & 90.51 & \textbf{90.20} & 90.42 
                                                           & 96.32 & 99.78 & 99.65 & 99.57 & 99.71 \\
        & SINCERE~\cite{Feeney2023sincere} & 78.72 & \textbf{89.91} & \textbf{90.62} & 90.18 & \textbf{90.58} 
                                           & 96.67 & \textbf{99.84} & \textbf{99.73} & \textbf{99.62} & \textbf{99.82} \\
        \cmidrule(lr){1-12}
        \multirow{6}{*}{ModelNet40} 
        & Cross-Entropy (baseline) & 32.57 & 33.60 & 55.45 & 78.85 & 55.64 
                                                      & 68.99 & 68.49 & 84.72 & 93.76 & 85.83 \\

        \cmidrule(lr){2-12}
         
        & InfoNCE~\cite{oord2018infonce} & 29.77 & \underline{66.43} & \underline{69.60} & \underline{71.98} & 75.19 
                                         & 60.54 & \underline{91.17} & \underline{92.57} & \underline{93.57} & \underline{94.58} \\
        & SimCLR~\cite{Chen2020simclr} & \underline{32.99} & 66.04 & 69.25 & 71.84 & \underline{75.37} 
                                       & \underline{64.09} & 90.72 & 92.53 & 93.32 & 94.56 \\
        
        \cmidrule(lr){2-12}
        
        & SupCon~\cite{khosla2020supcon} & 57.04 & 80.39 & 84.28 & 85.57 & 83.83 
                                         & \textbf{77.53} & 96.21 & 95.94 & 97.03 & 96.85 \\
        & $\epsilon$-SupInfoNCE~\cite{barbano2023unbiased} & \textbf{57.33} & 82.24 & \textbf{85.89} & \textbf{86.18} & \textbf{85.68} 
                                                           & 73.36 & \textbf{97.05} & \textbf{97.76} & 97.74 & \textbf{97.94} \\
        & SINCERE~\cite{Feeney2023sincere} & 56.78 & \textbf{82.27} & 85.60 & 85.93 & 85.63 
                                           & 74.35 & 96.99 & 97.42 & \textbf{97.80} & 97.72 \\
        \bottomrule
\end{tabular}
}
\end{table*}

\subsubsection{3D Shape Retrieval Performance}
\label{subsec:3d-retrieval}

Table~\ref{tab:all-metrics} reports the retrieval mAP of our best models compared with related works on the ModelNet dataset. Our approach ViT-B/16 with SINCERE loss, achieves the best mAP performance (95.5\%) on ModelNet10.

\begin{figure*}[htb]
    \centering
    \includegraphics[width=.99\linewidth]{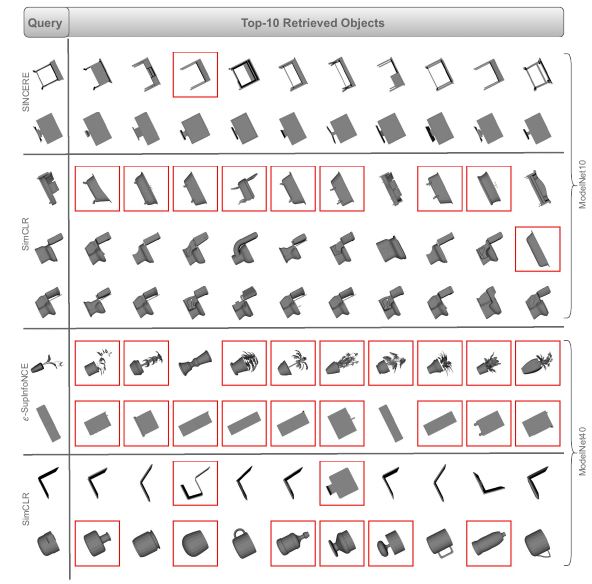}
    \caption{Qualitative comparison of Top-10 retrieved 3D objects using our best contrastive models on the ModelNet10 and ModelNet40 datasets. Each row shows retrieval results for a given model, top to bottom (SINCERE with ViT-B/16, SimCLR with DINO-B/16, $\epsilon$-SupInfoNCE with DINO-S/16, and SimCLR with DINO-B/16). For each query (leftmost column), the ten most similar objects are shown from left to right. Red boxes indicate incorrect class retrievals. SINCERE and $\epsilon$-SupInfoNCE yield more consistent and semantically accurate retrievals, highlighting their stronger representation learning capacity.}
    \label{fig:shape-retrieval}
\end{figure*}

In addition to some qualitative examples of objects retrieved in Figure~\ref{fig:shape-retrieval}, which help to understand the retrieval performance. Note that each row corresponds to a representative query (one view of an object from the test query set), with incorrectly retrieved shapes highlighted in red. 
For example, in the ModelNet40 dataset shown in Figure~\ref{fig:shape-retrieval}, when the query uses $\epsilon$-SupInfoNCE with the object ``plant'', the retrieval objects are confused because the model perceives the ``plant'' object as incorrect. In this specific case, this wrong correlation must be the specific value from the margin defined in this loss.

\begin{figure*}[htb]
    \centering
    \includegraphics[width=0.99\linewidth]{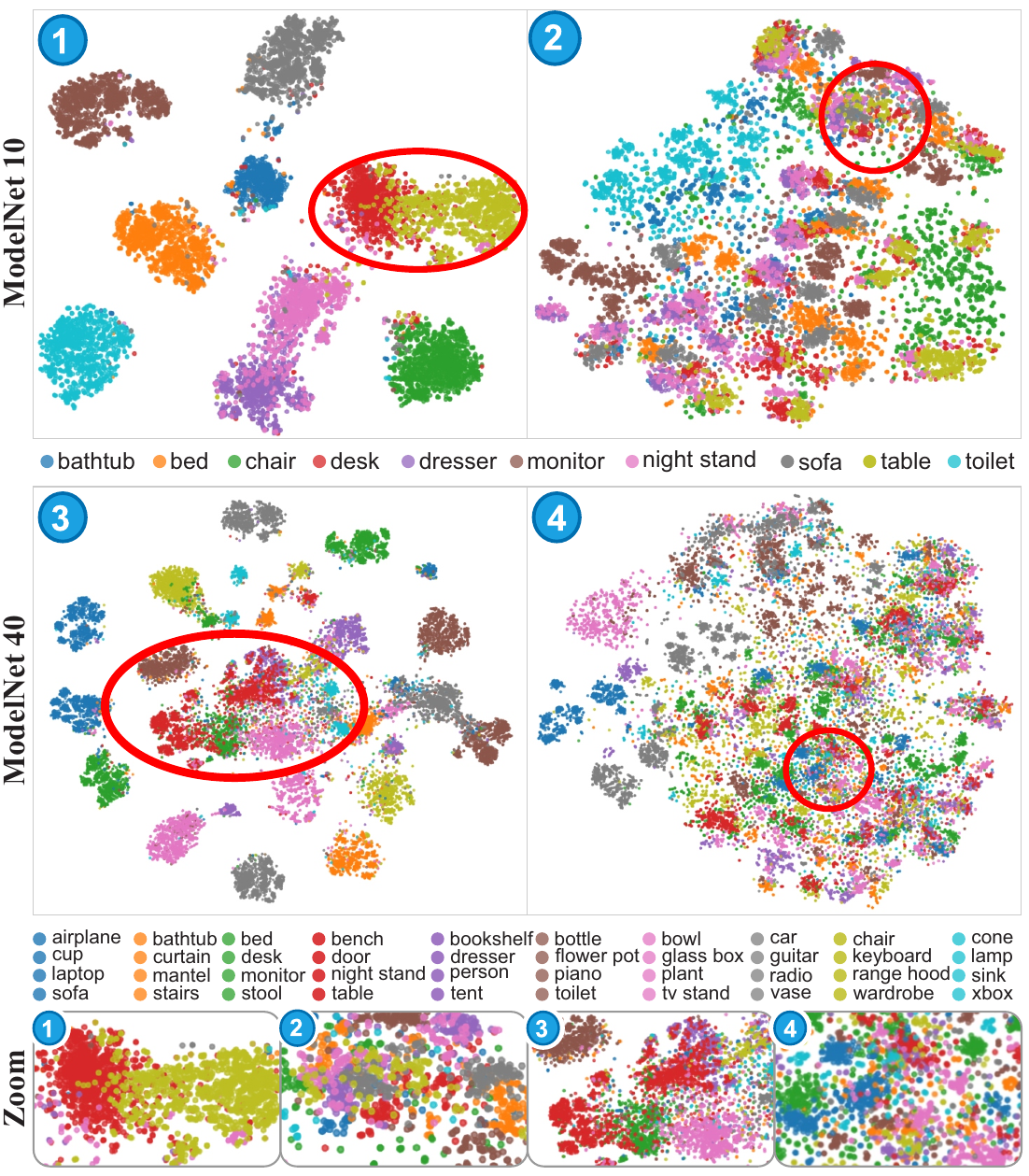}
   \caption{t-SNE visualization of the representation spaces obtained with the best contrastive learning settings after linear evaluation. Each row shows a number in the embedding space, which means: \circlemarker{1} SINCERE with ViT-B/16, \circlemarker{2} SimCLR with DINO-B/16, \circlemarker{3} $\epsilon$-SupInfoNCE with DINO-S/16, and \circlemarker{4} SimCLR with DINO-B/16. Red circles highlight areas with misclustered class overlap, and the zoomed-in views below offer a detailed look into such regions.}
  \label{fig:tsne}
\end{figure*}

To intuitively validate our best contrastive models, we explore the t-SNE~\cite{van2008visualizing} projection to visualize the behavior embeddings on the latent space for our approach, as shown in Figure~\ref{fig:tsne}. Both t-SNE projections from each dataset are shown in~\circlemarker{1} and~\circlemarker{2} to ModelNet10, \circlemarker{3} and~\circlemarker{4} to ModelNet40. 
For the ModelNet10 dataset, all ten categories are represented by dots of different colors, and the labels for each category/class are annotated at the bottom of the dataset. In contrast, in the ModelNet40 dataset, the colored dots are misaligned because there are just ten colors to represent 40 objects. Hence, in the same Figure~\ref{fig:tsne}, it's possible to see the zoomed regions with misclustered classes overlapped.

Figure \ref{fig:tsne} \circlemarker{1} and \circlemarker{3} show the effects when using supervised \acrshort{cl} to obtain a relatively clear separation (e.g., SINCERE with ViT-B/16 and $\epsilon$-SupInfoNCE with
DINO-S/16, respectively). Inevitably, our approach generates some grouped data points that are highlighted in a red circle. For example, in \circlemarker{1}, the classes ``table" and ``desk" are highly correlated semantically in terms of effects that relatively indistinguishable features of these shapes.
In contrast, under the self-supervised contrastive learning settings, as shown in \circlemarker{2} and \circlemarker{4}, the features have a more dispersed distribution, with mixed misclustering (e.g., SimCLR with DINO-B/16 on both datasets). The red circles in both cases demonstrate regions where shape features from different classes are highly correlated.

Furthermore, this intuitive analysis, which involves Figures \ref{fig:shape-retrieval} and \ref{fig:tsne}, highlights the limitations of both supervised and self-supervised learning using contrastive procedures in separating semantically similar classes or queries.

\subsection{RQ3: Comparison Performance with Related Works}
\label{subsec:results-related}

\begin{table*}[bht]
\centering
\renewcommand*{\arraystretch}{1.2}
\setlength{\tabcolsep}{4.0pt}
\caption{Evaluation with state-of-the-art methods. We report top-1 linear accuracy, $k$-NN ($k=10$) accuracy, mean Average Precision (mAP), and mAP@10. Top performing in \textbf{bold}.}
\label{tab:all-metrics}
\begin{adjustbox}{width=0.75\linewidth}
\begin{tabular}{lccccccc}
\toprule
Method & Backbone & Loss & \# of Views & Linear & $k$-NN & mAP & mAP@10 \\
\midrule

\textcolor{gray}{\textit{ModelNet10}} & & & & & & & \\

\acrshort{cmvlnet} \cite{peng2024contrastive}      & ResNet-18 & InfoNCE & 12 & 78.1 & 76.4 & -- & -- \\
\rowcolor{orange!10} \textbf{Ours}      & ViT-B/16  & SINCERE & 12 & \textbf{90.6} & \textbf{90.5} & \textbf{95.5} & \textbf{96.6} \\
\rowcolor{orange!10} \textbf{Ours}      & DINO-B/16 & SimCLR  & 12 & 78.8 & 86.6 & 51.4 & 90.8 \\

\midrule

\textcolor{gray}{\textit{ModelNet40}} & & & & & & & \\

MVCNN \cite{su2015multi}         & CNN       & --       & 12 & 89.9 & --   & 70.1 & -- \\
MVTN \cite{hamdi2021mvtn}        & CNN       & --       & 12 & 93.8 & --   & \textbf{92.9} & -- \\
EMVN \cite{esteves2019equivariant} & ResNet-18 & Triplet* & 12 & \textbf{94.5} & -- & 91.8 & -- \\
MVCL \cite{li2025multi}          & ResNet-50 & Triplet  & 12 & 83.5 & --   & --   & -- \\
\acrshort{cmvlnet} \cite{peng2024contrastive}    & ResNet-18 & InfoNCE  & 12 & 59.8 & 59.2 & --   & -- \\
\rowcolor{orange!10} \textbf{Ours}    & DINO-S/16 & $\epsilon$-SupInfoNCE & 12 & 86.2 & \textbf{87.0} & 58.7 & 87.3 \\
\rowcolor{orange!10} \textbf{Ours}    & DINO-B/16 & SimCLR & 12 & 75.4 & 84.5 & 50.7 & \textbf{89.3} \\
\bottomrule
\end{tabular}
\end{adjustbox}
\smallskip
\begin{minipage}{\linewidth}\centering
\footnotesize ``*'' denotes not contrastive method and ``-'' represents missing metric. 
\end{minipage}
\end{table*}

Last but not least, we answer RQ3 by comparing our proposed approach that combines ViT and contrastive learning with recent state-of-art methods.
We select state-of-art methods as mentioned in Section~\ref{sec:related} and presented in Table~\ref{tab:summary-relatedworks}. For this, we follow the same experimental setting as used in each method. We evaluate our proposed approach in the retrieval task and show \emph{mAP} and \emph{mAP@10} results.

Table \ref{tab:all-metrics} compares the experimental results carried out with 12 view-based shape objects. 
On ModelNet10, our best results are provided for our approach using SINCERE with ViT-B/16, surpassing the \acrshort{cmvlnet} \cite{peng2024contrastive} in 12.5\% \best and 14.1\% \best to linear evaluation and $k$-NN accuracy, respectively. The mAP metric performs well in comparison with our other method (SimCLR with DINO-B/16), showing an improvement in performance of about 44.1\% \best.
Now, on ModelNet40, EMVN \cite{esteves2019equivariant} and MVTN \cite{hamdi2021mvtn} obtained the best results for linear classification accuracy and mAP metric. Our approach ($\epsilon$-SupInfoNCE with DINO-S/16) obtained the best $k$-NN accuracy. 

\subsection{Discussion}\label{subsec:discussions}
With the quantitative results in mind, we will now focus on our guided research questions (RQ). 

To answer our first question (\textbf{RQ1}), we investigate the influence of end-to-end CNN and ViT backbones with CE loss in a scenario using only a 3D shape understanding pipeline. The results can be found in Table~\ref{tab:lineareval-acc-ce}. 
%
%
This evaluation shows the \emph{advantage} of using ViT-based backbones, such as ViT-S, ViT-B, DINO-S, or/and DINO-B architectures, over ResNet-50. The last, a CNN-based architecture, is one of the most used in the literature as shown in Section\,\ref{sec:related}, Table\,\ref{tab:summary-relatedworks}. This finding opens ways to explore vision transformers in the context of 3D multi-view data.

Our results for (\textbf{RQ2}) show that \textit{the choice of backbone, combined with a contrastive learning strategy, plays a critical role in the quality of fine-tuning and affects the features extracted from the backbone encoders.}
Regarding the effectiveness of contrastive learning, our results demonstrate that the recent contrastive losses (SINCERE and $\epsilon$-SupInfoNCE) overcome traditional approaches (SimCLR and SupCon, well established in the literature) in the proposed task. In addition, our experiments show which are the most promising combinations of ViT architectures and contrastive learning losses. In this way, our work proposes a benchmark for choosing such a combination for view-based 3D representation learning.
In addition, this finding shows how contrastive objectives can enhance feature learning from multi-view 3D data.

To enlarge our comprehension of RQ2, we show, as a qualitative result, the visualization of the feature space provided by backbones learned with different contrastive learning losses. The analysis of visual separation of different clusters of classes through t-SNE 2D projections and retrieval experiments reveals two different findings. First, we observe persistent challenges for all contrastive learning losses in distinguishing semantically similar categories of objects, such as tables and desks. This limitation manifests as overlapping clusters in the embedding space and occasional retrieval errors. This suggests that current methods may benefit from incorporating hierarchical or relational contrastive mechanisms to capture fine-grained shape differences better. This finding opens ways to apply different strategies of contrastive learning -- that we leave for future work. Second, supervised contrastive losses, mainly SINCERE, provide superior class separation compared to self-supervised approaches. Indeed, this confirms the main findings obtained by quantitative results in Section\,\ref{subsec:contrastive}, which also shows that supervised contrastive learning methods obtained the highest performance values for the evaluated datasets. 

For our third research question (\textbf{RQ3}), we proposed to unify both pipelines (contrastive and 3D shape understanding) with ViT backbones on the state-of-the-art methods. Our proposed method with a ViT-based backbone (ViT variants such as DINO) consistently surpasses state-of-the-art results for all compared metrics when evaluating the ModelNet10 dataset. Considering the ModelNet40 dataset, our proposed method shows an advance over state-of-the-art methods in terms of two metrics ($k$-NN and mAP@10), and a lower value in the other two (linear and mAP). 
First, a higher value for $k$-NN indicates that our proposed method is able to maintain local neighborhood preservation when combining ViT-based with contrastive learning losses. We argue that this advantage is mainly related to the transformer ability to capture local patterns in an image and, then, grouping images with similar patterns in the feature space through the contrastive learning loss. Second, a lower value for the linear metric shows that the proposed method is not able to learn a feature space that is linearly separable. In addition, this dataset (ModelNet40) shows to be more challenging when compared to other evaluated (ModelNet10).
We consistently see, even for CE loss, lower values in accuracy when compare both datasets (see Table~\ref{tab:lineareval-acc}). It is possible to see more mixture of different colors/classes in the 2D projection of the feature space (see Table~\ref{fig:tsne}, second row). This finding shows the benefit of using our proposed method, particularly when preserving the local neighborhood is important.

\section{\uppercase{Conclusions}}
\label{sec:conclusions}

Our findings challenge the long-standing dominance of CNNs in multi-view 3D analysis, demonstrating that ViT-based architectures paired with advanced contrastive objectives like SINCERE and $\epsilon$-SupInfoNCE can achieve relevant results (90.6\% accuracy, 95.5\% mAP on ModelNet10) while learning more geometrically meaningful embeddings. The success of our approach stems from its ability to simultaneously capture global shape semantics through ViTs' attention mechanisms and refine local discriminative features via contrastive optimization, a combination that proves particularly powerful for distinguishing structurally similar categories.

Beyond benchmark performance and the use of ViT-based backbones, our study reveals fundamental insights about representation learning itself. The persistent entanglement of semantically related classes (e.g., tables/desks) in embedding space, observed across both supervised and self-supervised paradigms, suggests an inherent limitation of current contrastive frameworks. This discovery opens new theoretical questions about how to better encode hierarchical shape relationships, a direction that require rethinking future research in 3D vision. 

Future progress may come from three directions highlighted by our analysis: (1) multi-scale ViT architectures that better preserve fine geometric details, (2) dynamic contrastive objectives that adapt to hierarchical class relationships, and (3) cross-modal distillation techniques to further bridge the synthetic-to-real gap. As 3D data becomes increasingly central to computer vision, the guidelines established here offers a strong foundation for future investigation.

\section*{\uppercase{Acknowledgements}}
This study was financed in part by the São Paulo Research Foundation (FAPESP -– grant 2024/09462-1, CNPq Fellowship (grant n.o 315158/2023-9) and Coordination for Higher Education Personnel Improvement (Finance Code 001 and grant 88887.969051/2024-00).

\bibliographystyle{apalike}
{\small
\bibliography{main}}



\end{document}